\documentclass[pdflatex,sn-mathphys-num]{sn-jnl}

\usepackage{graphicx}%
\usepackage{listings}
\usepackage{multirow}%
\usepackage{amsmath,amssymb,amsfonts}%
\usepackage{amsthm}%
\usepackage{mathrsfs}%
\usepackage[title]{appendix}%
\usepackage{xcolor}%
\usepackage{textcomp}%
\usepackage{manyfoot}%
\usepackage{booktabs}%
\usepackage{algorithm}%
\usepackage{algorithmicx}%
\usepackage{algpseudocode}%
\usepackage{hyperref}%
\usepackage{siunitx}%
\usepackage[nolist]{acronym}%

\usepackage{xcolor}
\usepackage{siunitx}
\usepackage[dash,dot]{dashundergaps}

\usepackage{soul}

\newcommand{\add}[1]{#1}
\newcommand{\rem}[1]{}

\begin{acronym}[]
    \acro{AI}{Artificial Intelligence}
    \acro{CA}{Cellular Automata}
    \acro{CNN}{Convolutional Neural Network}
    \acro{DoF}{Degrees of Freedom}
    \acro{DSP}{Digital Signal Processing}
    \acro{FPGA}{Field Programmable Gate Array}
    \acro{FPU}{Floating Point Unit}
    \acro{GI}{Gastro-Intestinal}
    \acro{LM}{Levenberg Marquardt}
    \acro{MCU}{Microcontroller Unit}
    \acro{MLP}{Multi-Layer Perceptron}
    \acro{MSE}{Mean-Squared Error}
    \acro{OR}{Operating Room}
    \acro{OTS}{Optical Tracking System}
    \acro{NCA}{Neural Cellular Automaton}
    \acrodefplural{NCA}{Neural Cellular Automata}
    \acro{PMT}{Permanent Magnet Tracking}
    \acro{PnO}{Position and Orientation}
    \acro{PSRAM}{Pseudo-Static RAM}
    \acro{SIMD}{Single Instruction Multiple Data}
    \acro{SLAM}{Simultaneous Localization and Mapping}
    \acro{SPI}{Serial Peripheral Interface}
    \acro{VO}{Visual Odometry}
    \acro{WCE}{Wireless Capsule Endoscopy}
\end{acronym}

\begin{document}

\title[eNCApsulate: NCA for Precision Diagnosis on Capsule Endoscopes]{eNCApsulate: Neural Cellular Automata for Precision Diagnosis on Capsule Endoscopes}


\author*[1]{\fnm{Henry John} \sur{Krumb}}\email{henry\_john.krumb@tu-darmstadt.de}

\author[1]{\fnm{Anirban} \sur{Mukhopadhyay}}

\affil*[1]{\orgdiv{Computer Science Department}, \orgname{TU Darmstadt}, \orgaddress{\street{Fraunhoferstr. 5}, \city{Darmstadt}, \postcode{64293}, \state{Hessen}, \country{Germany}}}


\abstract{\textbf{Purpose:}
\ac{WCE} is a non-invasive imaging method for the entire gastrointestinal tract, and is a pain-free alternative to traditional endoscopy.
It generates extensive video data that requires significant review time, and localizing the capsule after ingestion is a challenge.
Techniques like bleeding detection and depth estimation can help with localization of pathologies, but deep learning models are typically too large to run directly on the capsule.
\\
\textbf{Methods:} \acp{NCA} \rem{models} for bleeding segmentation and depth estimation are trained on capsule endoscopic images.
For monocular depth estimation, we distill a large foundation model into the lean \ac{NCA} architecture, by treating the outputs of the foundation model as pseudo ground truth.
We then port the trained \acp{NCA} to the ESP32 microcontroller, enabling efficient image processing on hardware as small as a camera capsule.
\\
\textbf{Results:} \acp{NCA} are\rem{29.1\%} more accurate (Dice) than other portable segmentation models, while requiring \add{more than 100x} fewer parameters stored in memory than other small-scale models.
The visual results of \acp{NCA} depth estimation look convincing, and in some cases beat the realism and detail of the pseudo ground truth.
Runtime optimizations on the ESP32-S3 accelerate the average inference speed significantly, by more than factor 3.
\\
\textbf{Conclusion:} With several algorithmic adjustments and distillation, it is possible to eNCApsulate \ac{NCA} models into microcontrollers that fit into wireless capsule endoscopes.
This is the first work that enables reliable bleeding segmentation and depth estimation on a miniaturized device, paving the way for precise diagnosis combined with visual odometry as a means of precise localization of the capsule -- on the capsule.}

\keywords{Wireless Capsule Endoscopy, Neural Cellular Automata, Edge AI, Image Segmentation, Depth Estimation}



\maketitle
\acresetall

\section{Introduction}\label{sec:intro}
\ac{WCE} is a non-invasive imaging modality that allows to record the \ac{GI} tract in its entirety, including the otherwise hard to reach small intestine.
This modality is compelling as it allows for pain-free and precise diagnosis, in a procedure that is not associated with stigma -- unlike traditional endoscopy.
However, capsule endoscopes collect hours of video data, which typically take 30 to 120 minutes of the physician's time to review~\cite{geClinicalApplicationWireless2003a}.
Video summarization and image classification are key technologies to counter this problem.
Performing the classification on the capsule directly further helps to significantly reduce the amount of data to be transmitted, as in most patients, only a small fraction of images shows pathological findings.

Another issue is to localize the camera capsule after it was swallowed, which still poses a challenge to researchers and engineers.
Knowing the exact location of the capsule allows for a targeted diagnosis, and assists to locate the capsule in cases of retention, which is the most significant complication associated with \ac{WCE}~\cite{liRetentionCapsuleEndoscope2008a}.
Several approaches discussed in the literature aim to include sophisticated sensors~\cite{ali2024implementing,vedaeiLocalizationMethodWireless2021} to accomplish this task, which require additional hardware to be built into the miniaturized capsule.
Others propose to track a permanent magnet inside the capsule using magnetic sensor arrays on the outside of the patient~\cite{caoRoboticWirelessCapsule2024}, requiring the patient to wear a sensor array for more than 12 hours while the capsule is travelling, which is uncomfortable and cumbersome.

Image-guided techniques like \ac{VO} are better suited for the task, as they enable to locate the capsule inside the \ac{GI} tract only based on available image data~\cite{ozyoruk2021endoslam}.
One key ingredient to \ac{VO} are depth images, which can be generated very well with contemporary monocular depth estimation models, even for a niche domain like \ac{WCE}~\cite{ozyoruk2021endoslam,hanDepthAnythingMedical2024,liAdvancingDepthAnything2024} which is underrepresented in training sets of foundation models.
\add{Even such monocular depth estimation models can be employed for \ac{VO}, as long as they predict absolute depth maps.}
However, all \ac{VO} approaches for \ac{WCE} are analyzed on PC hardware due to their high demands to compute, utilizing retrospective data collections.
As depth estimation models are too large for embedded platforms, precision navigation and diagnosis are not possible on the capsule.

Our vision is to
\rem{\textbf{bring Diagnosis and Visual Odometry onto the capsule endoscope}}
\add{bring Diagnosis and Visual Odometry onto the capsule endoscope}, enabling a localization of the capsule while it travels through the \ac{GI} tract without the requirement for additional sensing.
We start to pursue this vision by
\rem{\textbf{shrinking segmentation and depth estimation models to the size of a capsule-sized microcontroller}}
\add{shrinking segmentation and depth estimation models to the size of a capsule-sized microcontroller}, which would be impossible with contemporary \mbox{\ac{CNN}- or attention-based} models.
We employ \acp{NCA} models, which are a new family of bioinspired neural networks that are known to be not only accurate and robust, but also lightweight enough to fit on small scale hardware at a minimal memory footprint~\cite{kalkhof2023med}.

\begin{figure}
    \centering
    \includegraphics[width=1.0\linewidth]{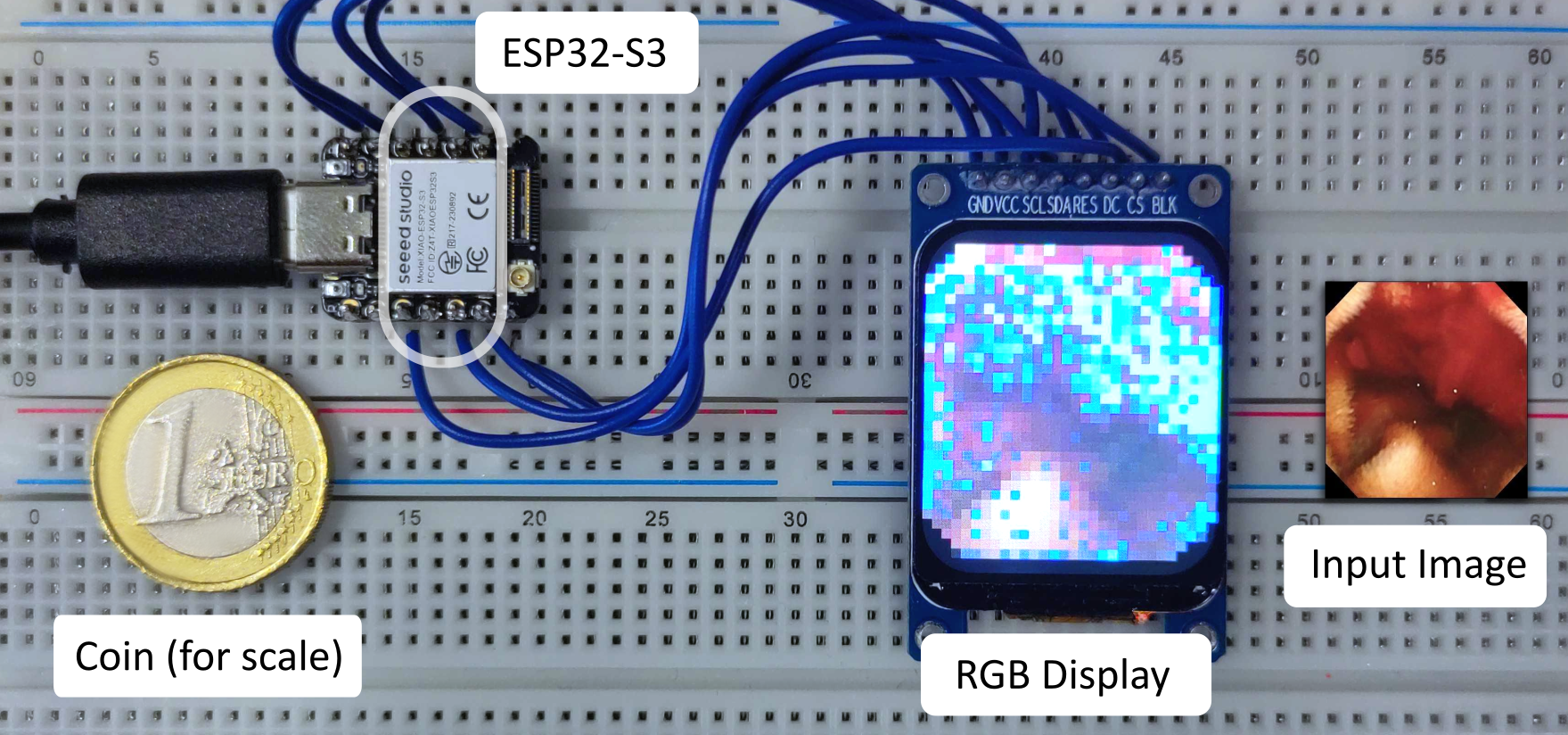}
    \caption{Experimental hardware setup for eNCApsulate, featuring a tiny version of the ESP32-S3 as the main component. The chip itself is miniaturized and can thus fit into a capsule endoscope (white contour overlay). An RGB display, connected directly to the chip, displays the segmentation of a capsule endoscopic image generated by an \ac{NCA}.}
    \label{fig:esp32setup}
\end{figure}

In this work, we present \textbf{eNCApsulate}, an approach for capsule endoscopic diagnosis and navigation that is lean enough to run on a capsule device itself.
This miniaturization is achieved by leveraging \ac{NCA}, an emerging lightweight and robust neural architecture, which we port to the ESP32 microcontroller.
We find that \ac{NCA} yield superior segmentations and convincing depth maps, while being 10x smaller than respective state-of-the-art models.
We evaluate their performance on an unseen dataset, presenting quantitative and qualitative results.

Our contributions are threefold:
\begin{enumerate}
    \item We are the first to introduce \ac{NCA} models to the field of capsule endoscopy, showing that \acp{NCA} predict accurate and convincing bleeding segmentations.
    Our segmentation is $29.1\%$ more accurate (Dice) than the predictions of other small-scale segmentation models.
    
    \item Our work is the first to distill a large foundation model into a small \ac{NCA}.
    We investigate the task of depth estimation, finding that \ac{NCA} produce convincing depth maps from RGB images by leveraging pseudo ground truth from Depth Anything V2.

    \item We demonstrate that our proposed approach can be ported to a microcontroller, such that the segmentation and depth estimation are carried out on the capsule itself.
    The distilled segmentation and depth estimation models are ported to an ESP32-S3 microcontroller, which is small enough to fit inside a common capsule endoscope.
    This is an initial step towards precise pathology diagnosis, combined with localization via \ac{VO} on the capsule.
\end{enumerate}

\section{Related Work}\label{sec:related}
\textbf{WCE Bleeding Detection:}
The main research challenge in bleeding detection on \ac{WCE} data is the scarcity of annotated data for this rather exotic modality.
Even though available datasets promise a high quantity of images, they comprise mostly redundant healthy samples and only few bleeding images.
The largest \ac{WCE} dataset, KvasirCapsule~\cite{smedsrud2021kvasir}, contains 47,238 annotated images, of which 446 show blood -- these 446 images stem from only two individual patients.
Vats et al. \cite{vatsLabelsPriorsCapsule2022} address the issue of data scarcity by inducing domain priors into a contrastive learning scheme for bleeding detection, utilizing multiple datasets.
\add{}

\noindent\textbf{WCE Depth Estimation:} 
With EndoSLAM, Ozyoruk et al. propose a holistic pipeline for depth estimation and \ac{VO} for \ac{WCE}, introducing a public dataset of multiple anatomies and capsules~\cite{ozyoruk2021endoslam}, in which depth maps were acquired in a simulated setting.
Obtaining ground truth depth maps for monocular depth estimation in \ac{WCE} is a challenge that is being actively investigated.
In their recent study, Jeong et al. leverage simulated data to generate labelled ground truth, and then use a CycleGAN-based approach to bridge the sim-to-real gap~\cite{jeongDepthEstimationMonocular2024}.
Universal depth estimators like the Depth Anything Model are also investigated~\cite{hanDepthAnythingMedical2024,liAdvancingDepthAnything2024} for their applicability in endoscopic imaging and reconstruction.
\add{Although the more recent DepthAnythingV2 has not yet applied to \ac{WCE} images, these studies show that there is an increased demand for monocular depth estimation on \ac{WCE} images, and contemporary depth estimators achieve promising results.}

\noindent\textbf{Miniaturized Models on Capsule Endoscopes:} Bringing advanced CNN-based models to the minimal capsule hardware is also a recent research challenge.
Sahafi et al. pioneered this field by proposing a custom capsule design, featuring a Kendryte K210 chip, allowing to run simple CNN models on the capsule for the purpose of polyp classification~\cite{sahafi2022edge}.
However, simple CNN models do not scale equally for every downstream task, and have a larger memory footprint for tasks like depth estimation or segmentation\add{, prohibiting the deployment on the capsule}.

\noindent\textbf{NCA on Minimal Hardware:} Portable architectures like \acp{NCA} promise to bring advanced neural image processing for universal downstream tasks to miniaturized hardware.
For instance, Kalkhof et al. successfully ported \acp{NCA} to the Raspberry Pi~\cite{kalkhof2023med,kalkhof2023m3d} and Smartphones~\cite{kalkhofUnsupervisedTrainingNeural2024}.\rem{<newline>}
In this work, we leverage \acp{NCA} to generate bleeding segmentation and depth images on hardware that is tiny enough to fit in a capsule endoscope.
With our framework, we are able to generate convincing depth maps at a minimal memory footprint, which is otherwise impossible with contemporary CNN-type or transformer architectures.

\section{Methods}\label{sec:methods}
We \rem{will} start by explaining the general idea behind \ac{NCA} training and inference, and \rem{will} continue by outlining the eNCApsulate models with all of their tweaks.
Finally, we will elaborate on the experimental design used in this study to train NCA models on the PC and transfer them to the ESP32 microcontroller platform.

\subsection{Neural Cellular Automata}

\begin{figure}
    \centering
    \includegraphics[width=1.0\linewidth]{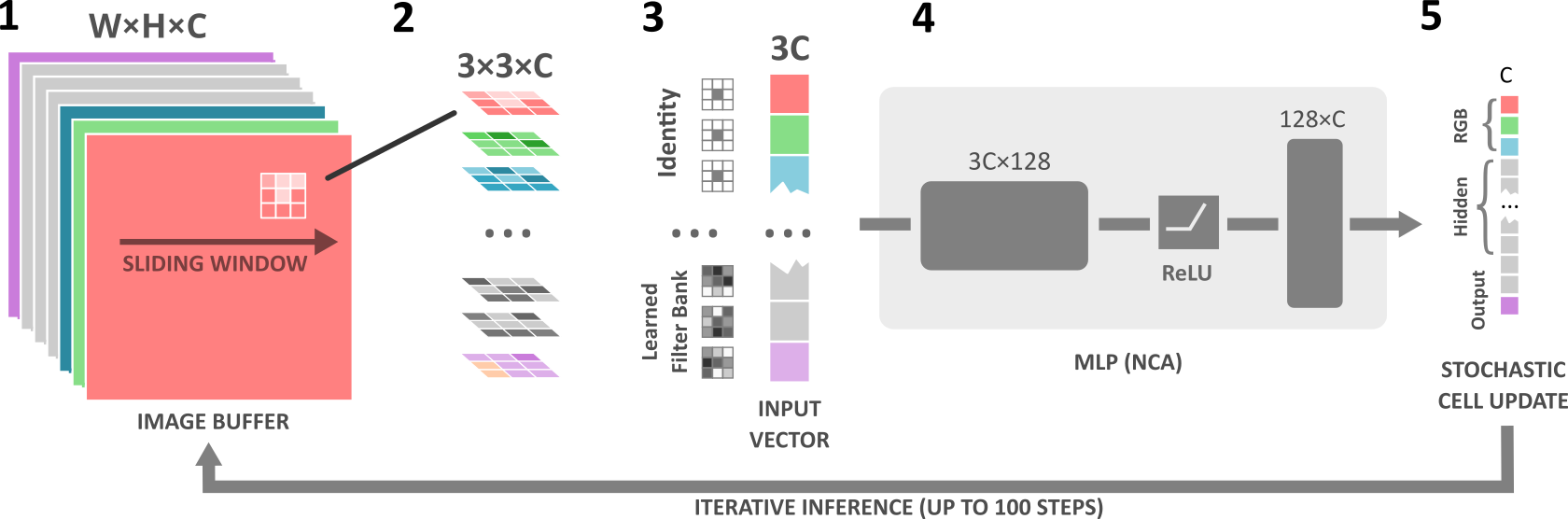}
    \caption{eNCApsulate architecture for lightweight segmentation or depth estimation. (1) The channels of the input RGB image are augmented to the match the input + hidden + output channel dimension $C$. (2) The input image is then processed by a learned bank of $3\times 3$ filters, and for each pixel, the concatenated result (3) is fed as an input to the NCA MLP network (4). The NCA MLP computes the image update for each cell. The result is an update vector (5) that is added to the input image buffer by a chance of 50\% (stochastic cell update).}
    \label{fig:architecture}
\end{figure}

\acp{NCA} are an emerging family of neural models that are gaining traction in the application of medical image processing.
Their working principle combines the ideas of two bio-inspired systems, bringing together neural networks and cellular automata.
\acp{NCA} work on an image grid with an extended channel dimension, where a common local rule is applied to each of the cells in an iterative fashion.
Figure~\ref{fig:architecture} illustrates the operation that is performed on each cell, in each timestep: First, the Moore neighborhood of each cell is aggregated by applying $3 \times 3$ image filters.
We use three filter banks, one consisting of identity filters which only yield the current cell's state, and two learned filter banks that aggregate the cell's neighborhoods using a different filter matrix in each channel.
The scalar result for each channel is stored in a vector that is passed to an \ac{MLP}, representing the common rule applied to every cell.
The final result is then added to the original image buffer once all new cell states are computed.
However, only \rem{50\%}\add{80\%} of cells are stochastically updated in each timestep to relax the simultaneous cell updates on the entire grid.
For a more detailed overview of the \ac{NCA} architecture and its key ideas, we point the interested reader to the comprehensive "Growing NCA" paper by Mordvintsev et al.~\cite{mordvintsev2020growing}.

\subsection{NCA Model Architecture and Training}

We train two models, namely \mbox{\textbf{eNCApsulateS}} for segmentation and \mbox{\textbf{eNCApsulateD}} for depth estimation.
eNCApsulateS operates on an 18 channel image, whereas eNCApsulateD uses 22 channels total.
In both models, the first three channels are fixed, as these are the RGB channels necessary to store the image data.
In the last channel, the \ac{NCA} produces the segmentation or depth map output respectively.
All channels in between are hidden channels that the \ac{NCA} model uses to retain information between individual time steps.
The hidden channels and output channel are initialized noise, as we found that this increases the robustness of the training.

eNCApsulateD is trained on a \rem{small} subset of the KID2~\cite{pmid28580415} dataset, which is passed through Depth Anything Model V2~\cite{depth_anything_v2} in order to obtain pseudo ground truth depth maps.
The resulting depth maps are \rem{carefully} automatically curated, as we could not fully trust the foundation model and hence had to remove \rem{430} image samples for which the generated depth maps appear flat.
To determine if a depth map is flat, we leverage its normalized gradient magnitude and accept it if it exceeds a threshold of $1.1$.
After applying this strategy, 727 annotated samples remained, which were used to train the depth estimator.

eNCApsulateD is trained with a combination of three losses: Mean Squared Error (MSE), Structural Similarity (SSIM) loss and an image gradient loss, weighed $\lambda_{MSE} = 1.0$, $\lambda_{SSIM} = 1.0$ and $\lambda_{grad} = 0.1$ respectively.
During training, we make use of batch duplication as this has improved training stability in prior work~\cite{kalkhof2023med}.
\rem{Our m}Minibatches have size 8 (duplicated: 16) and are comprised of \rem{\mbox{$100 \times 100$} random image patches that we obtained from the} \add{cropped} capsule endoscopic samples \rem{by cropping}, which are \rem{then} downsampled to \mbox{$64 \times 64$} patches.
\add{By resizing},\rem{This way,} we can cut down the high VRAM requirements during training\rem{ -- assuming that no excessive global knowledge is required to predict an accurate depth map}.
\rem{We exploit the finding that \acp{NCA} are very robust against changes in image scale~\cite{kalkhof2023med}.}

For the qualitative evaluation of eNCApsulateD, we select a subset of the KvasirCapsule dataset~\cite{smedsrud2021kvasir} with interesting benchmark images, which we sort into five categories, each of which contains five sample images: Blood, Bubbles, Complex Folds, Debris and Foreign Body.

All models are implemented in Python with Pytorch, and trained on a PC equipped with an NVidia GeForce GTX 3090.

\subsection{Porting NCAs to Microcontrollers}

Once the \ac{NCA} model is trained and properly tested on PC hardware, the next step is to port eNCApsulate to the ESP32-S3 microcontroller.
As shown in Figure~\ref{fig:esp32setup}, we use a tiny variant of the ESP32-S3 microcontroller in our experimental setup.
Despite not being a dedicated hardware accelerator for neural networks, the ESP32-S3 offers several functionalities that allow us to run \acp{NCA} efficiently.
Most importantly, it features \ac{SIMD} instructions for instruction-level parallelism.
These are especially useful for matrix operations, which are needed for the forward pass in the \ac{NCA}'s \ac{MLP}.
We make use of \ac{SIMD} instructions whenever applicable, increasing the average runtime speed for a single inference from \SI{9}{\second} to \SI{3}{\second}.
The ESP32-S3 also features a proper \ac{FPU}, significantly accelerating floating point instructions required for the depthwise convolution.

Since we cannot rely on code optimizations under the hood of a framework like PyTorch, the entire 
inference loop is implemented from scratch in ANSI C.
A major difference in our NCA implementation on the microcontroller is in the order in which the inference steps are executed.
The stochastic cell update is the first step to be executed, as it is a 50:50 condition for the rest of the code to run through.
After that, we compute the filter operations of the depth-wise convolution; however, we do not store the results of these filters in separate buffer matrices.
Instead, we only make use of two buffers: One which is the actual image buffer, and the update buffer which is added to the image after all cell updates were computed.
We will provide our implementation upon acceptance of this paper, so that these optimization steps can be traced easily.

\subsection{Accelerating Inference on the ESP32-S3}

Since inference in \ac{NCA} models is an iterative process, they trade model size for runtime.
Although this property allows us to bring the model to lightweight architectures, it comes with the price of a rather slow inference process.
Typically, an \ac{NCA} needs around 100 time steps (forward passes) to converge, which is negligible for inference on the GPU, but costly on the microcontroller, where each time step takes roughly \SI{65}{\milli\second}.

We therefore add an extension to eNCApsulateS in the form of a temporal regularization scheme to reduce unnecessary time steps, in order to reduce the average inference time.
In particular, we interrupt the inference process after a certain number of steps if no significant change in the hidden channels is observed.
Once a minimum number of steps (10) is reached, we take the total absolute difference between each two consecutive hidden channel tensors, sum them up for all cells and normalize them by the number of entries.
If this absolute difference falls below a threshold (we use 0.1), a cooldown counter will be decremented from 5 to 0.
Once it reaches 0, the inference is stopped at the current time step, otherwise it is reset to 5.
A full description of this algorithm can be found in our public supplemental material.

\section{\add{Experimental }Results}\label{sec:results}
\begin{figure}[t]
    \centering
    \includegraphics[width=1.0\linewidth]{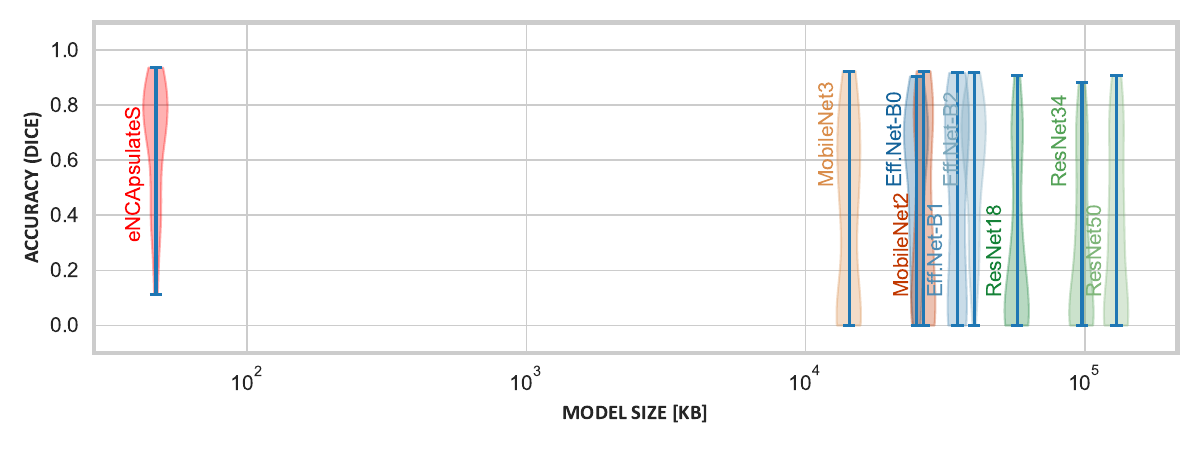}
    \caption{Accuracy of different lightweight segmentation models (blue) vs. eNCApsulateS (green), and their model size in kilobytes, on a logarithmic scale.}
    \label{fig:accuracyvsparameters}
\end{figure}

\begin{table}
    \centering
    \begin{tabular}{l|cc|r}
        \toprule
        Model &             Dice 	$\uparrow$ &   IoU $\uparrow$ & Size [\rem{k}B] $\downarrow$\\
        \midrule
        \add{
        nnUNet          & $0.582$            & $0.462$         &   268,113,106 \\
        \hline
        EfficientNet-B0 & $0.465 \pm 0.298$  & $0.353 \pm 0.259$ &  25,005,876 \\
        EfficientNet-B1 & $0.400 \pm 0.338$  & $0.310 \pm 0.289$ &  35,028,420 \\
        EfficientNet-B2 & $0.509 \pm 0.288$  & $0.389 \pm 0.254$ &  40,185,164 \\
        MobileNetV3s &    $0.350 \pm 0.330$  & $0.267 \pm 0.276$ &  14,342,596 \\
        MobileNetV2 &     $0.345 \pm 0.333$  & $0.265 \pm 0.285$ &  26,515,780 \\
        ResNet18 &        $0.251 \pm 0.298$  & $0.184 \pm 0.240$ &  57,312,836 \\
        ResNet34 &        $0.254 \pm 0.291$  & $0.184 \pm 0.231$ &  97,745,476 \\
        ResNet50 &        $0.295 \pm 0.313$  & $0.220 \pm 0.257$ &  130,084,420\\
        \midrule
        eNCApsulateS (Ours) & $0.576 \pm 0.291$ & $0.460 \pm 0.270$ & \textbf{47,152} \\
        }
        \rem{EfficientNet-B0 &          0.631 &          0.461 &        1562.87\\
        EfficientNet-B1 &          0.687 &          0.523 &          2189.28\\
        EfficientNet-B2 &          0.586 &          0.414 &        2511.57\\
        MobileNetV2 &          0.598 &          0.427 &        1657.24\\
        ResNet18 &          0.739 &          0.587 &        3582.05\\
        ResNet34 &          0.571 &          0.400 &      6109.09\\
        ResNet50 &          0.035 &          0.018 &         8130.28\\
        MobileNetV3s &          0.390 &          0.421 &      896.41\\
        \midrule
        eNCApsulateS (Ours) & \textbf{0.791} & \textbf{0.655} & \textbf{44.32} \\
        }
        \bottomrule
    \end{tabular}
    \caption{Comparison of different small-scale segmentation models (backbones for U-Net) and eNCApsulateS, evaluated on \add{a held-out testset, which is a subset of} the KID2 dataset. Results are computed \rem{for}\add{by an ensemble of models trained on the 5-fold-split.}\rem{ each individual image and are then averaged.}}
    \label{tab:accuracy}
\end{table}

\begin{figure}
    \centering
    \includegraphics[width=1.0\linewidth]{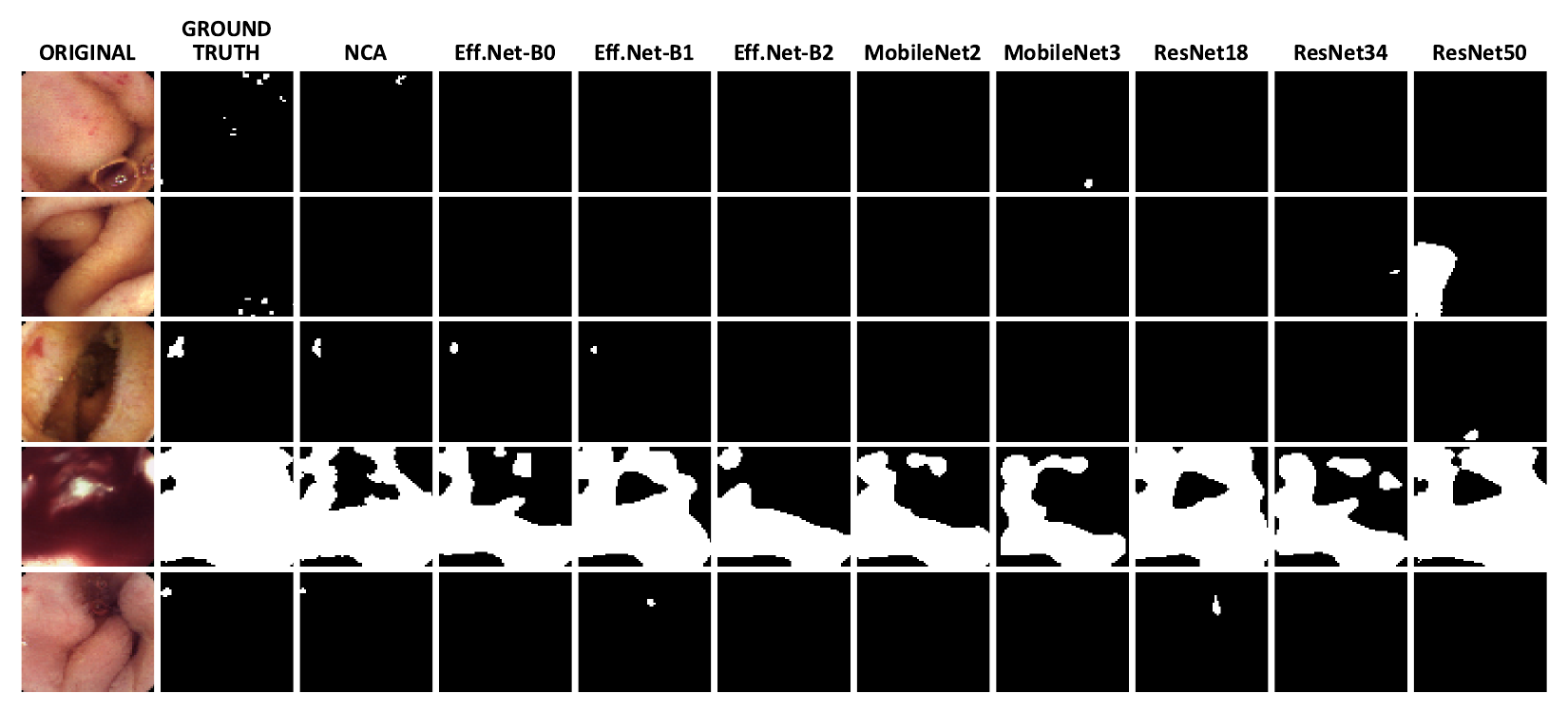}
    \caption{Qualitative segmentation results for eNCApsulateS, compared to other lightweight segmentation models based on CNN.}
    \label{fig:segcomparison}
\end{figure}

\begin{figure}
    \centering
    \includegraphics[width=0.8\linewidth]{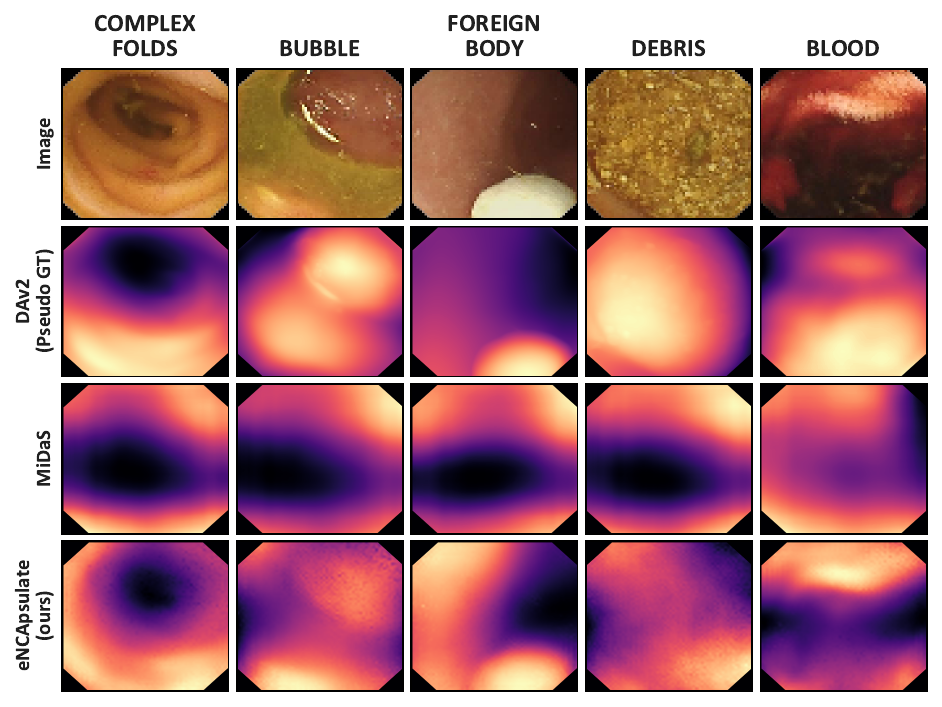}
    \caption{Visual comparison of different monocular depth estimation approaches on a part of the benchmark dataset (subset of KvasirCapsule). eNCApsulate was trained on the KID2 dataset, whereas the other models are foundation models.}
    \label{fig:depthcomparison}
\end{figure}

\begin{figure}
    \centering
    \includegraphics[width=0.7\linewidth]{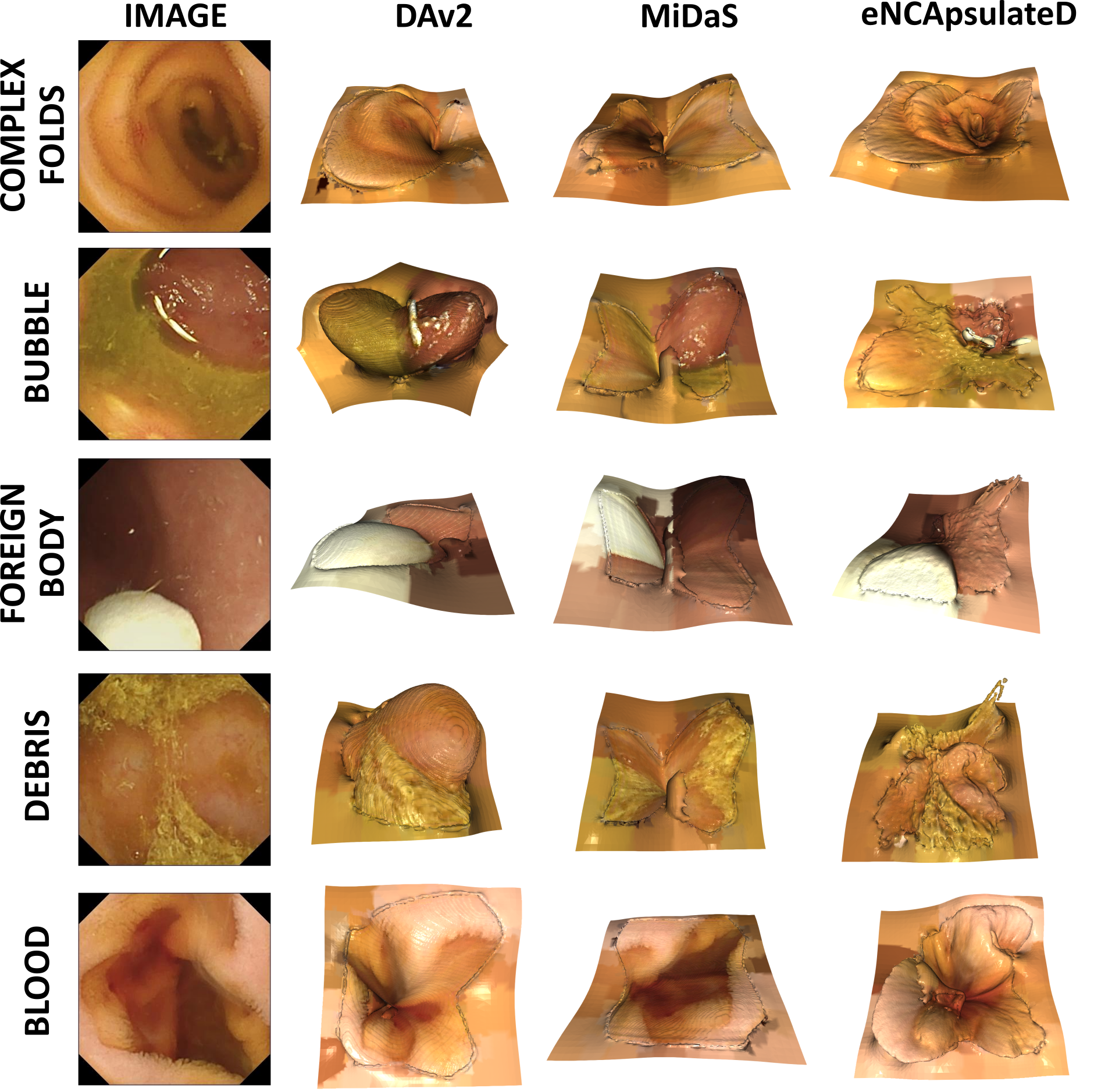}
    \caption{3D Projections of generated RGBD images with baselines and eNCApsulateD, for the five categories of our baseline set.}
    \label{fig:renderings}
\end{figure}

\noindent\textbf{Segmentation with eNCApsulateS:} Our segmentation approach is compared to various lightweight backbones for U-Net, which can also be miniaturized to run on confined hardware architectures.
\add{All baselines were pre-trained on ImageNet, with their encoders being frozen before re-training.}
\add{To estimate the upper bound of what is achievable, we also compare against nnUNet which is the current SOTA for medical image segmentation.}
All baseline models were trained on the same dataset and \add{five-fold} split as eNCApsulateS for comparison.
\add{We also used similar settings and hyperparameters for the baseline training (1000 epochs with early stopping, Dice BCE loss, batch size 4) and also used the same image pre-processing.}
eNCApsulateS clearly outperforms the other lightweight segmentation models that are optimized for size, while being 10x smaller than the smallest baseline (MobileNet V3s) as it is shown in Table \ref{tab:accuracy}.
\rem{At the same time, the \ac{NCA}-based model is $29.1\%$ more accurate than the best performing baseline (EfficientNet-B1) in terms of Dice score.}
Figure~\ref{fig:segcomparison} shows that eNCApsulateS performs well on images with different sizes of bleedings, even if they are tiny (bottom row) or excessively large (4th row from above).
In some cases, we found that the ground truth labels of KID2 are rather coarse and often oversegment the bleeding \add{which explains the poor upper-bound performance, even for strong models like nnUNet}.
Even in such cases, eNCApsulateS turned out to be robust as it only delineates the actual bleedings\add{, being nearly on-par with nnUNet}.

\noindent\textbf{Depth Estimation with eNCApsulateD:} NCAs for depth estimation are data efficient and generalize well on an unseen dataset.
Reconstructed 3D volumes are demonstrated in Fig.~\ref{fig:renderings}.
One central limitation of our approach is the absence of a proper ground truth for depth, forcing us to use the best-performing foundation model as a pseudo ground truth.
However, qualitative results (Fig.~\ref{fig:depthcomparison}) indicate that eNCApsulateD performs well, and in some cases produces more convincing depth maps than the foundation models.
We attribute this phenomenon to two effects: Firstly, the foundation models were trained on a multitude of mostly real-world datasets.
Medical data are underrepresented in such datasets, (capsule-)endoscopic data is even less likely to be found.
In the end, all models struggle with challenging objects such as bubbles or debris.
Secondly, \ac{NCA} have proven earlier to be robust against data shifts and inconsistencies in training data~\cite{kalkhof2023med}, and they are data-efficient thanks to their common local update rule.
We assume that they perform well in the segmentation and depth estimation task thanks to these properties, even though the training datasets were rather small.

\noindent\textbf{Temporal Regularization:} Our temporal regularization approach works well for capsule endoscopic videos, as most frames in capsule endoscopic videos show healthy findings and only few are clinically relevant.
In the case of the KvasirCapsule video showing obvious bleedings, our temporal regularization strategy needs 1,222,998 NCA steps total (at the same segmentation quality), whereas 6,988,560 NCA steps are needed without the early cut-off.
On the ESP32-S3, this reduction by factor 5 means on average reduction of inference time from \SI{3}{\second} to less than \SI{1}{\second} per image.
As most capsules record frames at 2-3 FPS, such an improvement in runtime speed at similar accuracy implies that accurate bleeding segmentation on the capsule can be performed in reasonable time that aligns with the frequency of recorded images.

\section{Conclusion}\label{sec:conclusion}
In this paper, we have successfully ported \acp{NCA} to the ESP32 microcontroller platform for segmentation and depth estimation on a chip as small as a capsule endoscope.
While such tasks typically require large models with sizes of several megabytes, we manage to train models of less than 70~kB that produce convincing segmentations and depth maps.
Our segmentation model eNCApsulateS segments the bleedings more accurately than U-Net based small-scale models, and manages to perform a full image segmentation in less than three seconds.
Inference can be further accelerated by stopping the inference process at low hidden channel activity without losing precision.
The depth estimator eNCApsulateD predicts depth maps with realistic appearance even in difficult cases (e.g. folds of the colon, dark areas, bubbles), making \acp{NCA} a promising candidate technology in confined, but also in less confined settings.
However, a hurdle that is yet to overcome is the overall data scarcity and the quality of ground truth in the field of \ac{WCE}, which can be mitigated by an increase of shared data repositories and advancements of simulation environments like VR-Caps~\cite{incetan2020vrcaps}.
In the future, we plan to integrate the depth estimation approach with similarly lean feature extraction methods to attempt a localization of the capsule camera.
To accomplish this goal, absolute depth estimation is necessary, which is already incorporated in models like DepthAnythingV2 and EndoSfMLearner and will be further investigated for eNCApsulateD.
Our work paves the way for such a localization strategy, hopefully enabling sensor-less navigation of capsules within the \ac{GI} tract.


\section*{Declarations}

\textbf{Funding:} This work is partially supported by Norwegian Research Council project number 322600 Capsnetwork.\\
\textbf{Code availability:} Our code is available on GitHub: \url{https://github.com/MECLabTUDA/eNCApsulate}\\
\textbf{Supplementary information:} We will publish a supplemental material file along with the manuscript.\\
\textbf{Data availability:} KvasirCapsule is available under a public license. KID2 is distributed under a custom academic license by request.\\
\textbf{Conflict of Interest:} The authors have no conflict of interest to declare.
\\
\\
Other declarations are not applicable.

\bibliography{bibliography}

\end{document}